%% file: neurips_2025.tex
\documentclass{article}

\PassOptionsToPackage{numbers, compress}{natbib}
\usepackage[main, final]{neurips_2025}

\usepackage[utf8]{inputenc} % allow utf-8 input
\usepackage[T1]{fontenc}    % use 8-bit T1 fonts
\usepackage{hyperref}       % hyperlinks
\usepackage{url}            % simple URL typesetting
\usepackage{booktabs}       % professional-quality tables
\usepackage{amsfonts}       % blackboard math symbols
\usepackage{nicefrac}       % compact symbols for 1/2, etc.
\usepackage{microtype}      % microtypography
\usepackage{xcolor}         % colors

\newcommand{\td}[1]{\textcolor{black}{#1}}
\usepackage{pifont}
\usepackage{graphicx}
\usepackage{subfigure}
\usepackage{amsmath}
\usepackage{algorithm}
\usepackage{algorithmic}
\usepackage{amssymb}

\title{Multi-Agent Reinforcement Learning with Communication-Constrained Priors}

\author{%
Guang Yang$^{1}$ \quad Tianpei Yang$^{12*}$ \quad Jingwen Qiao$^{2}$ \quad Yanqing Wu$^{2}$ \quad Jing Huo$^1$ \\ 
\textbf{Xingguo Chen}$^4$ \quad \textbf{Yang Gao}$^{123}$\thanks{Corresponding Author: Tianpei Yang; Yang Gao} \\
$^1$State Key Laboratory for Novel Software Technology, Nanjing University \\
$^2$School of Intelligence Science and Technology, Nanjing University\\
$^3$School of Network Security and Information Technology, YiLi Normal University\\
$^4$Nanjing University of Posts and Telecommunications\\
\texttt{\{yangg,jingwenqiao,yanqingwu\}@smail.nju.edu.cn}\\
\texttt{\{tianpei.yang,huojing,gaoy\}@nju.edu.cn}\\
\texttt{chenxg@njupt.edu.cn}
}
\usepackage{multirow}
\usepackage{array}
\usepackage{siunitx}
\usepackage{makecell}
\newcolumntype{M}[1]{>{\centering\arraybackslash}m{#1}}

\begin{document}

\maketitle

\begin{abstract}
Communication is one of the effective means to improve the learning of cooperative policy in multi-agent systems. However, in most real-world scenarios, lossy communication is a prevalent issue. Existing multi-agent reinforcement learning with communication, due to their limited scalability and robustness, struggles to apply to complex and dynamic real-world environments. To address these challenges, we propose a generalized communication-constrained model to uniformly characterize communication conditions across different scenarios. Based on this, we utilize it as a learning prior to distinguish between lossy and lossless messages for specific scenarios. Additionally, we decouple the impact of lossy and lossless messages on distributed decision-making, drawing on a dual mutual information estimatior, and introduce a communication-constrained multi-agent reinforcement learning framework, quantifying the impact of communication messages into the global reward. Finally, we validate the effectiveness of our approach across several communication-constrained benchmarks.
  % The abstract paragraph should be indented \nicefrac{1}{2}~inch (3~picas) on
  % both the left- and right-hand margins. Use 10~point type, with a vertical
  % spacing (leading) of 11~points.  The word \textbf{Abstract} must be centered,
  % bold, and in point size 12. Two line spaces precede the abstract. The abstract
  % must be limited to one paragraph.
\end{abstract}

\input{main/intro}
\input{main/rela}
\input{main/prob}
\input{main/meth}
\input{main/expri}
\input{main/conc}

\begin{ack}
This work is supported in part by National Natural Science Foundation of China (62192783, 62276128, 62506157, 62276142), Jiangsu Natural Science Foundation (BK20243051), Jiangsu Science and Technology Major Project (BG2024031), the Fundamental Research Funds for the Central Universities (14380128, KG202514), Nanjing International/Hong Kong, Macao and Taiwan Science and Technology Cooperation Plan (202308031) and the Collaborative Innovation Center of Novel Software Technology and Industrialization.
% Use unnumbered first level headings for the acknowledgments. All acknowledgments
% go at the end of the paper before the list of references. Moreover, you are required to declare
% funding (financial activities supporting the submitted work) and competing interests (related financial activities outside the submitted work).
% More information about this disclosure can be found at: \url{https://neurips.cc/Conferences/2024/PaperInformation/FundingDisclosure}.

% Do {\bf not} include this section in the anonymized submission, only in the final paper. You can use the \texttt{ack} environment provided in the style file to automatically hide this section in the anonymized submission.
\end{ack}

% \section*{References}

% References follow the acknowledgments in the camera-ready paper. Use unnumbered first-level heading for
% the references. Any choice of citation style is acceptable as long as you are
% consistent. It is permissible to reduce the font size to \verb+small+ (9 point)
% when listing the references.
% Note that the Reference section does not count towards the page limit.
% \medskip

% {
% \small

% [1] Alexander, J.A.\ \& Mozer, M.C.\ (1995) Template-based algorithms for
% connectionist rule extraction. In G.\ Tesauro, D.S.\ Touretzky and T.K.\ Leen
% (eds.), {\it Advances in Neural Information Processing Systems 7},
% pp.\ 609--616. Cambridge, MA: MIT Press.

% [2] Bower, J.M.\ \& Beeman, D.\ (1995) {\it The Book of GENESIS: Exploring
%   Realistic Neural Models with the GEneral NEural SImulation System.}  New York:
% TELOS/Springer--Verlag.

% [3] Hasselmo, M.E., Schnell, E.\ \& Barkai, E.\ (1995) Dynamics of learning and
% recall at excitatory recurrent synapses and cholinergic modulation in rat
% hippocampal region CA3. {\it Journal of Neuroscience} {\bf 15}(7):5249-5262.
% }

% plainnat

\bibliographystyle{plainnat}
\bibliography{references}

%%%%%%%%%%%%%%%%%%%%%%%%%%%%%%%%%%%%%%%%%%%%%%%%%%%%%%%%%%%%
\newpage
\appendix

\section{Appendices}
\subsection{Details on Experiement}

\subsubsection{Environmental  Description}
% 本研究的实验评估在广泛使用的多智能体粒子环境中进行，具体选取了以下三个具有代表性的MPE场景：
The experimental evaluation of this study was conducted in a widely used multi-agent particle environment(MPE). Specifically, the following three representative MPE scenarios were selected:

% \Simple_Tag：此场景模拟了经典的捕食者-猎物追逐博弈。$N$个具有有限视野半径的捕食者（本实验中$N$设置为3，并在附录中补充了6个及9个智能体的评估结果）需要协作捕捉一个猎物。捕食者均由独立的强化学习策略控制，具备更大的体型、较低的移动速度以及较小的加速度。猎物的行为策略是预定义的（随机采样100个侯选位置，在其感知范围内，通过距离评估函数选择最优移动方向）。当任意捕食者接触到猎物时，所有捕食者均会获得正奖励。此外环境中还设置了2个不可移动障碍物，增加了追逐任务的复杂性。
\textbf{Simple\_Tag}: This scenario simulates the classic predator-prey pursuit game. $N$ predators with limited field of view radius ($N$ is set to 3 in this experiment, and the evaluation results of 6 and 9 agents are supplemented in the appendix) need to collaborate to capture a prey. The predators are controlled by independent reinforcement learning strategies, with larger body size, lower movement speed and smaller acceleration. The behavior strategy of the prey is predefined (randomly sampling 100 candidate positions, selecting the optimal movement direction within its perception range through the distance evaluation function). When any predator contacts the prey, all predators will receive a positive reward. In addition, 2 immovable obstacles are set in the environment to increase the complexity of the pursuit task. 

% Simple\_Spread：此场景要求$N$个智能体（本实验中固定为3个）通过协作覆盖$N$个预先设定的固定地标。与Simple\_Tag类似，每个智能体的视野范围有限。智能体的奖励是所有智能体到其最近未被覆盖地标的最小距离之和的负值，此外当智能体间发生碰撞时会被惩罚。因此智能体既要快速移动以覆盖各自的目标地标，又要避免相互碰撞，并确保所有地标都被唯一覆盖。
\textbf{Simple\_Spread}: This scenario requires $N$ agents (fixed to 3 in this experiment) to cover $N$ pre-set fixed landmarks through collaboration. Similar to Simple\_Tag, each agent has a limited field of view. The reward for the agent is the negative of the sum of the minimum distances of all agents to their nearest uncovered landmarks. In addition, when there is a collision between agents, they will be penalized. Therefore, the agents must move quickly to cover their respective target landmarks, avoid collisions with each other, and ensure that all landmarks are uniquely covered.

% Simple\_Reference：此场景是一个独特的协作导航任务，场景包括2个智能体和3个不同颜色的固定地标。与前两个场景不同的是，每个智能体都具有全局观测能力，其挑战性在于：每个智能体的目标地标仅能通过另一个智能体的通信消息得知。集体奖励取决于到达目标地标的智能体数量。
\textbf{Simple\_Reference}: This scenario is a unique collaborative navigation task, which includes 2 agents and 3 fixed landmarks of different colors. Unlike the previous two scenarios, each agent has global observation capabilities. The challenge is that the target landmark of each agent can only be known through the communication messages of another agent. The collective reward depends on the number of agents that reach the target landmark. 

\subsubsection{Parameters Setting}
% 每个智能体的演员网络是具有两个隐藏层的神经网络，每个隐藏层具有64个神经元，使用ReLU激活，输出层使用tanh激活函数，输出动作。所有智能体共享一个中心化的评论家网络，其隐藏层结构与演员网络类似。JSD网络，通信消息和动作分别通过单层32神经元的编码器，使用Jensen-Shannon散度估计互信息下界。CLUB网络，中间层有32个神经元，使用ReLU激活，输出层使用tanh激活，建模有损消息与动作的条件分布。
In this experiment, an NVIDIA RTXA5000 24GB GPU was used. The actor network of each agent is a neural network with two hidden layers, each with 64 neurons, activated with ReLU, and the output layer with tanh activation function to output actions. All agents share a centralized critic network, whose hidden layer structure is similar to the actor network. In the JSD network, communication messages and actions are passed through a single-layer encoder with 32 neurons, respectively, and the mutual information lower bound is estimated using Jensen-Shannon divergence. In the CLUB network, the middle layer has 32 neurons, activated with ReLU, and the output layer uses tanh activation to model the conditional distribution of lossy messages and actions.

% 所有网络均使用Adam优化器，演员网络、JSD网络、CLUB网络学习率为$1\times10^{-4}$，评论家网络学习率为$1\times10^{-3}$，折扣因子设置为0.95，目标网络更新率设置为0.01。经验回放池大小为$1\times10^{5}$，消息缓冲区大小为$1\times10^{3}$，批处理大小通常为1024（在处理Simple\_Tag任务中6个及9个智能体的场景时，为了避免CC-MADDPG的训练过程超出GPU内存的问题，批处理大小调整为512）。随机种子设置为1。每个回合的时间步固定为25步。所有模型均训练$4.0\times10^{6}$个总时间步。当总时间步超过1024步后，每100个总时间步进行一次模型参数更新。为了进行更为有效的动作探索，在训练初期为演员网络的输出动作添加了Ornstein-Uhlenbeck噪声，其参数为$\theta=0.15$和$\sigma=0.2$，噪声尺度会在训练初期随训练回合数线性衰减。
Adam optimizer is used for all networks. The learning rate of actor network, JSD network and CLUB network is $1\times10^{-4}$, the learning rate of critic network is $1\times10^{-3}$, the discount factor is set to 0.95, and the target network update rate is set to 0.01. The replay buffer size is $1\times10^{5}$, the message buffer size is $1\times10^{3}$, and the batch size is usually 1024 (in the Simple\_Tag task, when the number of agents is 6 and 9, the batch size is adjusted to 512 to avoid the problem of CC-MADDPG training process exceeding GPU memory). The random seed is set to 1. The time step of each round is fixed to 25 steps. The total time step of training for all models is $4.0\times10^{6}$. When the total time steps exceed 1024, the model parameters are updated every 100 total time steps. For more effective action exploration, Ornstein-Uhlenbeck noise is added to the output actions of the actor network at the beginning of training with parameters $\theta=0.15$ and $\sigma=0.2$. At the beginning of training, the noise scale decays linearly with the number of training rounds.

% 评估时加载训练完成的模型，在各种测试环境下分别运行100个回合，每个回合同训练过程固定为25个时间步，主要记录和比较各算法的平均回合累计奖励以及标准差。
During the evaluation, the trained model is loaded and run for 100 episodes in various test environments. The training process of each episode is fixed to 25 time steps. The average episode cumulative reward and standard deviation of each algorithm are mainly recorded and compared.

\subsection{More Experiemental Analysis}
% 本研究还进一步探讨了当智能体数量增加时各算法的可扩展性。在Simple\_Tag场景中，理论上，由于捕食者之间没有碰撞惩罚，那么增加捕食者的数量应该能够直接提升捕获概率，获得更高的累计奖励。但是实验结果（见表格）表明，智能体数量的增加同时也显著提升了任务的协作复杂度，这对所有被测算法都构成了挑战。
This study further explored the scalability of each algorithm when the number of agents increases. In the Simple\_Tag scenario, theoretically, since there is no collision penalty between predators, increasing the number of predators should be able to directly increase the probability of capture and obtain higher cumulative rewards. However, the experimental results (see table \ref{tab:scalability_analysis}) show that the increase in the number of agents also significantly increases the collaborative complexity of the task, which poses a challenge to all tested algorithms. 

\begin{table}[!htbp]
\centering
\caption{Performance of Multi-Agent Algorithms under Varying Number of Agents and Communication Constraints}
\label{tab:scalability_analysis}
\footnotesize
\setlength{\tabcolsep}{2pt}
\resizebox{1.0\columnwidth}{!}{
\begin{tabular}{@{}llccccccc@{}}
\toprule
\multirow{2}{*}{Task Scenario} & \multirow{2}{*}{Testing Environment} & \multicolumn{6}{c}{Algorithms} \\
\cmidrule(lr){3-8}
 & &  \makecell{FC-MADDPG} & \multicolumn{3}{c}{\makecell{Dropout-MADDPG}} & \makecell{MADDPG} & \makecell{CC-MADDPG} \\
\cmidrule(lr){4-6}
 & & & 0.2 & 0.5 & 0.8 & & \\
\midrule

\multirow{8}{*}{\shortstack{Simple\_Tag\\(3 agents)}}
& Unrestricted & 75.9$\pm$65.3 & 70.3$\pm$65.9 & 65.9$\pm$62.0 & 72.1$\pm$66.1 & 5.2$\pm$12.8 & \textbf{134.7}$\pm$89.9 \\
& Light MBC (3) & 72.9$\pm$65.3 & 70.9$\pm$65.9 & 65.9$\pm$62.1 & 72.5$\pm$66.1 & 5.2$\pm$12.8 & \textbf{133.6}$\pm$89.9 \\
& Medium MBC (6) & 67.2$\pm$53.7 & 71.0$\pm$65.7 & 67.1$\pm$62.3 & 72.1$\pm$66.1 & 5.5$\pm$15.6 & \textbf{134.9}$\pm$90.9 \\
& Heavy MBC (8) & 54.4$\pm$43.8 & 69.8$\pm$66.5 & 67.9$\pm$63.2 & 72.7$\pm$66.0 & 6.0$\pm$14.2 & \textbf{131.4}$\pm$86.2 \\
& Light DBC (5) & 19.5$\pm$38.2 & 69.0$\pm$65.8 & 67.3$\pm$62.4 & 70.8$\pm$64.2 & 23.5$\pm$41.5 & \textbf{136.9}$\pm$89.7 \\
& Medium DBC (3) & 10.9$\pm$26.5 & 70.1$\pm$65.8 & 67.8$\pm$62.4 & 70.7$\pm$63.8 & 44.8$\pm$58.4 & \textbf{135.3}$\pm$85.1 \\
& Heavy DBC (1) & 1.5$\pm$5.2 & 68.7$\pm$64.3 & 66.3$\pm$60.5 & 71.4$\pm$65.7 & 71.2$\pm$70.9 & \textbf{138.0}$\pm$88.1 \\

\midrule

\multirow{7}{*}{\shortstack{Simple\_Tag\\(6 agents)}}
& Unrestricted & 138.5$\pm$88.0 & 84.2$\pm$70.3 & 78.5$\pm$68.1 & 73.1$\pm$61.6 & 6.2$\pm$12.8 & \textbf{131.8}$\pm$89.9 \\
& Light MBC (3) & 123.9$\pm$77.0 & 83.5$\pm$70.3 & 78.1$\pm$67.3 & 72.9$\pm$61.3 & 6.2$\pm$13.1 & \textbf{131.9}$\pm$90.3 \\
& Medium MBC (6) & 78.5$\pm$63.1 & 82.1$\pm$70.9 & 79.6$\pm$67.9 & 72.0$\pm$64.3 & 6.4$\pm$12.8 & \textbf{133.2}$\pm$84.3 \\
& Heavy MBC (8) & 69.8$\pm$59.5 & 81.3$\pm$69.0 & 78.6$\pm$65.4 & 75.2$\pm$61.8 & 7.0$\pm$15.3 & \textbf{131.8}$\pm$85.8 \\
& Light DBC (5) & 14.7$\pm$31.0 & 83.0$\pm$71.0 & 80.1$\pm$68.6 & 75.7$\pm$62.5 & 6.6$\pm$11.8 & \textbf{135.3}$\pm$89.4 \\
& Medium DBC (3) & 5.4$\pm$13.4 & 75.5$\pm$66.3 & 79.3$\pm$64.4 & 77.2$\pm$63.5 & 8.0$\pm$14.5 & \textbf{127.4}$\pm$89.1 \\
& Heavy DBC (1) & 3.8$\pm$10.0 & 79.1$\pm$65.3 & 72.2$\pm$69.3 & 76.5$\pm$63.7 & 11.1$\pm$20.9 & \textbf{122.1}$\pm$77.8 \\

\midrule

\multirow{7}{*}{\shortstack{Simple\_Tag\\(9 agents)}}
& Unrestricted & 83.4$\pm$72.5 & 77.7$\pm$71.5 & 68.0$\pm$61.0 & 19.8$\pm$27.7 & 7.2$\pm$15.4 & \textbf{78.2}$\pm$78.1 \\
& Light MBC (3) & 77.9$\pm$69.5 & 77.8$\pm$71.9 & 68.1$\pm$61.1 & 19.5$\pm$30.4 & 7.3$\pm$15.4 & \textbf{77.7}$\pm$72.6 \\
& Medium MBC (6) & 50.3$\pm$64.5 & 77.0$\pm$71.7 & 66.3$\pm$60.9 & 19.0$\pm$25.5 & 8.3$\pm$12.7 & \textbf{78.9}$\pm$72.5 \\
& Heavy MBC (8) & 38.9$\pm$35.3 & 77.4$\pm$72.5 & 67.3$\pm$61.9 & 16.5$\pm$28.0 & 8.1$\pm$17.0 & \textbf{77.6}$\pm$73.4 \\
& Light DBC (5) & 9.7$\pm$18.6 & 42.9$\pm$53.2 & 68.9$\pm$64.6 & 19.9$\pm$36.7 & 13.9$\pm$22.5 & \textbf{82.7}$\pm$74.4 \\
& Medium DBC (3) & 8.0$\pm$16.5 & 27.1$\pm$41.6 & 56.3$\pm$60.5 & 20.4$\pm$33.7 & 18.1$\pm$32.5 & \textbf{83.7}$\pm$71.0 \\
& Heavy DBC (1) & 5.8$\pm$10.7 & 21.0$\pm$33.4 & 49.0$\pm$52.7 & 21.1$\pm$34.9 & 17.5$\pm$30.8 & \textbf{83.2}$\pm$70.6 \\

\bottomrule
% \vspace{-2.0em}
\end{tabular}
}
% \vspace{-1.0em}
\end{table}

% 具体来看，FC-MADDPG虽然在6个智能体的理想环境下取得了高达138.5的平均奖励，但是其性能在引入约束条件后或者智能体数量再增多时均出现明显下降，再次证明了其对理想通信的依赖性和在复杂协调下的局限性。对于Dropout-MADDPG系列算法，在智能体数量增加时其鲁棒性优势减弱，难以应对高协作复杂度和强通信约束并存的环境，比如在9个智能体的场景中，dropout-0.2在基于距离判定的丢包模式下，其平均奖励分别降低至42.9、27.1和21.0。
Specifically, although FC-MADDPG achieved an average reward of up to 138.5 in an ideal environment with 6 agents, its performance dropped significantly after the introduction of constraints or when the number of agents increased, once again proving its dependence on ideal communication and its limitations under complex coordination. For the Dropout-MADDPG series of algorithms, its robustness advantage weakens when the number of agents increases, and it is difficult to cope with environments with high collaborative complexity and strong communication constraints. For example, in the scenario of 9 agents, the average rewards of dropout-0.2 in distance-based constraint are reduced to 42.9, 27.1 and 21.0 respectively. 

% 相比之下，CC-MADDPG展现出更强的鲁棒性与可扩展性。在6个智能体和9个智能体的场景中，其在理想环境下的效果接近FC-MADDPG，而在引入通信约束后更是始终保持领先的性能水平。其次，在智能体数量增加时，其相对于简单Dropout方法的鲁棒性优势在受限环境中更为明显。比如在9个智能体的场景中，在短距离阈值约束下，CC-MADDPG能取得83.2的平均奖励，而此时dropout-0.5只有49.0。这说明CC-MADDPG在处理更复杂的通信约束协作问题时，具有比简单消息丢失方法更强的鲁棒性与有效性。
In contrast, CC-MADDPG exhibits stronger robustness and scalability. In the scenarios of 6 and 9 agents, its performance in an ideal environment is close to that of FC-MADDPG, and it always maintains a leading performance level after the introduction of communication constraints. Secondly, when the number of agents increases, its robustness advantage over the simple message dropout method is more obvious in a constrained environment. For example, in the scenario of 9 agents, under the heavy distance-based constraint, CC-MADDPG can achieve an average reward of 83.2, while dropout-0.5 is only 49.0. This shows that CC-MADDPG is more robust and effective than the simple message dropout method when dealing with more complex communication-constrained problems.

\end{document}

%% file: main/intro.tex
\section{Introduction}
\label{Intro}

In multi-agent reinforcement learning (MARL) with partial observations, collaboration poses a significant challenge \cite{oroojlooy2023review}. Communication is one of the effective measures to improve the learning of cooperative policy in MARL \cite{zhu2024survey}, widely applied in scenarios such as autonomous driving \cite{chen2024communication, xu2024distributed} and cooperative drones \cite{chafii2023emergent, kim2024cooperative}. 
However, real-world scenarios are far from ideal and communication between agents often faces various constraints, specifically: (1) limited communication bandwidth, meaning only a limited amount of message can be transmitted, and (2) lossy communication, where transmitted message may be subject to interference, delay, loss, and other issues.

\td{Much of the research on communication-constrained multi-agent reinforcement learning (MARL) focuses on the limited bandwidth issue \cite{wang2019learning, lee2021learning, hu2024learning, zhou2024semantic}. In this setting, it is assumed that the communication channel is ideal, which means that transmission is real-time and lossless. Therefore, these methods typically only need to focus on how to effectively allocate communication resources (e.g., communication bandwidth, communication medium) to promote cooperation among agents. For example, compressing the communication information to extract and transmit parts beneficial to cooperation \cite{liu2020multi}, and systematically allocating communication mediums to avoid competition, among other problems \cite{kim2019learning}. }

\td{However, in most real-world scenarios, communication links are uncertain, and lossy communication is more prevalent. 
In this setting, existing work typically addresses two types of issue: noise interference and communication delay. 
Approaches to addressing noise interference typically involve modeling the unknown noise distribution and constructing learnable processes to adaptively optimize cooperative policy \cite{freed2020communication}.  
Another more common issue is communication delay, which frequently occurs in wireless network environments, which refers to the non-real-time  transmission of message, It has been shown to impact the performance of multi-agent behaviors \cite{zhang2020succinct, yuan2023dacom}.
Approaches to addressing communication delay focus primarily on how to remove the impact of delayed message on cooperative policy, such as constructing communication buffer \cite{zhang2020succinct} and determining when to communicate \cite{yuan2023dacom}.
Nevertheless, the above methods are all based on ideal assumptions about communication delay, and may not be applicable to more complex and unknown real-world scenarios such as underwater and caves. The reasons are as follows: (1) These methods \textit{lack scalability due to the lack of consideration of the common characteristics of lossy communication in different unknown scenarios}. (2) These methods \textit{lack robustness due to the lack of consideration of the dilemma of promoting the relevance of effective communication messages and suppressing the relevance of lossy communication information}.}
% \begin{itemize}
%     \item These methods lack scalability due to the lack of consideration of the common characteristics of communication delay in different unknown scenarios.
%     \item These methods lack robustness due to the lack of consideration of the dilemma of ``promoting the relevance of effective communication information'' and ``suppressing the relevance of lossy communication information''.
% \end{itemize}

\td{In order to overcome the above challenges, we first propose a generalized model of lossy communication to uniformly characterize the communication conditions in different scenarios, such as underwater, caves, and wireless networks. Based on this model, for specific scenarios, we use the lossy communication model as a learning prior to differentiate between lossy and lossless messages. Furthermore, for the second issue, we decouple the impact of different types of messages on distributed decision-making,  borrowed from dual mutual information estimatior. On one hand, we enhance the positive impact of lossless messages on decision-making by maximizing the lower bound of mutual information. On the other hand, we reduce the negative impact of lossy messages on decision-making by minimizing the upper bound of mutual information. Finally, we propose a communication-constrained MARL framework and validate its robust performance in two communication-constrained scenarios serving as benchmarks: Markov Model-Based and Distance-Based communication constraints.}

%% file: main/rela.tex
\section{Related Work}
The challenge of multi-agent collaboration within communication-constrained environments has undergone extensive scholarly investigation. Existing research predominantly addresses two key dimensions: one concerns the optimization of interactions in bandwidth-constrained scenarios, and the other focuses on enhancing system robustness under conditions of message loss. 

\subsection{MARL with Communication Bandwidth Constraints}
Research on communication bandwidth constraints has primarily focused on how to achieve efficient interaction among multiple agents under limited bandwidth resources.
\citet{zhang2019efficient} proposed the VBC method, which reduces communication overhead by constraining the variance of information exchanged between agents during the training phase, thereby eliminating noise from the messages. \citet{kim2019learning} addressed the issue of medium contention in the information transmission process, proposing the SchedNet framework. By incorporating a MAC protocol from the wireless communication field, this framework alleviates bandwidth pressure and addresses the scheduling problem of constrained agents. \citet{mao2020learning} introduced the Gated-ACML algorithm, which uses a gating mechanism to filter messages. It adaptively prunes unnecessary information by determining whether a message is beneficial using a threshold, thus reducing the number of messages exchanged. \citet{hu2023event} proposed the ETCNet framework, which converts limited bandwidth into a penalty threshold for event-triggered strategies, improving communication efficiency in multi-agent systems by sending messages only when necessary. 
% 这些方法本质上是对通信信息进行有效的压缩，这和信息瓶颈思想不谋而合，即剔除任务无关的冗余信息，提取任务相关的精炼信息。这些压缩方法已经被验证在相似任务上具有一定泛化能力。但是压缩并不等效于鲁棒，在不稳定通信或通信有损的环境下，这些压缩的方法未必有效，例如，不稳定通信就极易丢失重要信息，导致策略的质量大幅度下降。

These methods essentially perform effective compression on messages, which aligns closely with the information bottleneck \cite{tishby2015deep} principle—eliminating task-irrelevant redundant information and extracting task-relevant refined information. Such compression approaches have been validated to exhibit a certain degree of generalization capability in similar tasks \cite{yang2025learning}. However, compression is not equivalent to robustness \cite{yann2020learning, wang2022pac}. In environments with unstable or lossy communication, these compressed methods may not be effective. For instance, unstable communication is highly prone to losing critical messages, leading to a significant degradation in the quality of policies.

\subsection{MARL with Lossy Communication}
Research on lossy communication has primarily focused on the robustness of multi-agent systems under non-ideal communication conditions, such as environmental noise, transmission delays, and packet loss. \citet{freed2020communication} proposed a novel differentiable communication method using a randomized message encoding scheme, where it mathematically equates discrete communication channels to simulated channels with additive noise, enabling gradient backpropagation through these channels. \citet{kim2019message} proposed the Message-Dropout training method, based on the concept of Dropout, which randomly drops communication messages from other agents during training to improve robustness against communication errors during execution. \citet{zhang2020succinct} introduced the TMC method, which leverages temporal locality to reduce redundant messages and incorporates a message buffering mechanism to enhance robustness against packet loss, making it suitable for bandwidth-constrained and packet-loss-prone network environments. \citet{yuan2023dacom} proposed DACOM, an adaptive delay-aware multi-agent communication model in which agents can learn to schedule waiting times for messages from other agents, which is suitable for delay-sensitive tasks and high-latency scenarios. 
% These research outcomes provide a diverse range of solutions to address the complexities of communication environments.

Nevertheless, the aforementioned studies predominantly remain confined to communication-constrained problems in specific scenarios. While these research approaches help to analyze particular issues in depth, its conclusions may not fully apply to other more complex and dynamic real-world application scenarios. To better align with real-world scenarios, in the next section, we have further refined the formalization of MARL with communication constraints.

% 为了更加符合现实场景，在下一章中我们对通信受限下多智能体强化学习的形式化做了进一步完善。

%% file: main/prob.tex
\section{Problem Formulation}
\label{pf}
% \subsection{Decentralized Partially Observable Markov Dcecision Process}
Considering a fully collaborative multi-agent task where each agent is in a partially observable and communicative environment, it can be modeled as a decentralized partially observable Markov decision process (Dec-POMDP) \cite{albrecht2024multi} with communication, which is defined by the following $10$-tuple, $\mathcal{G}=\langle \mathcal{N}, \mathcal{S}, \mathcal{O}, \mathcal{A}, \mathcal{T}, \mathcal{R}, \mathcal{Z}, \mathcal{M}, \mathbb{I}, \gamma \rangle$. $\mathcal{N}=\{1,\dots, N\}$ is denoted by the set of $N$ agents. 
$\mathcal{S}$ is denoted by the set of environmental states. 
$\mathcal{O}=\{o^i\}_{i=1}^N$ is observation set of all agents, and $o^i$ is an observation set for agent $i$, which is determined by observation function $\mathcal{Z}(s, i)$. 
$\mathcal{A}=\mathcal{A}^1 \times\dots\times\mathcal{A}^N$ is the set of agents' joint action space, where $\mathcal{A}^i$ is denoted by the action space of agent $i$.
$r=\mathcal{R}(s,a)$ is the global reward signal shared by the agents.
$\mathcal{T}:\mathcal{S}\times\mathcal{A}\times\mathcal{S}\to[0, 1]$ is denoted by the transition probability. 
$\gamma \in [0, 1)$ is the discount factor. 

Furthermore, 
$\mathcal{M}$ is denoted by the message space, and $m^{ij}\in \mathcal{M}$ is denoted by the message sent by agent $i$ to agent $j$. 
To uniformly characterize the communication conditions in different environments, we further introduce a notation $\mathbb{I}=\{\iota^{ij}\}_{i\neq j}$, where $\iota^{ij}\in \{0, 1\}$ is denoted by the communication link status when agent $i$ sends message to agent $j$, where $1$ indicates effective communication, while $0$ indicates lossy communication. 
The set of message received by the agent $i$ can be defined by $M^i=\{\iota^{ji}m^{ji}\}_{j=1}^{j\neq i}$. In this case where the communication link is dynamic, the information received by each agent is different.
% 在这个通讯链路受限情况下，每个智能体收到的信息是具有差异的。

Then, given observation $o^i_t$ and message $M^i$ at the time step $t$, each agent $i$ uses a stochastic policy $\pi^i(\cdot|o^i, M^i)$ to choose actions. We denote the joint policy as $\pi=\{\pi^1,\pi^2,\cdots,\pi^N\}\in \Pi$, where $\Pi$ is the joint policy space. In cooperative MARL, the collaborative team aims to ﬁnd a joint policy to maximize the total expected discounted return $J(\pi)=\mathbb{E}_{\pi}\left[\sum_{t=0}^\infty \gamma^t r_t\right]$.

%% file: main/meth.tex
\section{Robust Learning with Communication-Constrained Priors}
%  为了overcome 有损通信下多智能体策略学习的挑战，这一节引入了我们的算法框架，其中具体包括，通讯受限先验建模以此来捕捉通讯链路的动态变化，信息的行为影响估计以此刻画不同通讯信息与智能体行为的相关性，最后提出通讯受限下的多智能体强化学习方法，优化提升不同通讯环境下的策略学习鲁棒性
To overcome two challenges of multi-agent collaborative policy learning with lossy communication, this section will introduce our algorithmic framework, which specifically includes: (1) modeling communication-constrained priors to capture the dynamics of communication links; (2) estimating messages' behavioral impacts to characterize the correlation between different communication messages and agent behaviors; and (3) proposing a communication-constrained MARL approach to optimize and enhance the robustness of policy learning across diverse communication environments.

\subsection{Communication-Constrained Priors Modeling}
% 在通讯受限的多智能体强化学习中，对于未知的场景下通信受限建模至关重要且存在挑战，其中，不仅需要尽可能抽象出影响多智能体策略学习的共性问题，而且还需要在不同现实场景下可泛化，因此，我们首先提出了一个binary的通讯链路参数刻画信息可靠性，此外，针对不同场景，通讯链路可进一步形式化为：
In communication-constrained MARL, modeling constrainted communication in unknown scenarios is crucial and challenging. It is necessary not only to abstract the common problems that affect multi-agent policy learning as much as possible, but also to generalize them in different real-world scenarios. Therefore, we first propose a binary communication link parameter $\iota$ to characterize message reliability.
In addition, for different scenarios, the communication link can be further formalized as follows,
\begin{equation}
    \iota^{ij}= f_{\theta_e}(s^{ij}),
\end{equation}
where $\theta_e$ is the parameter determined by the environment, and $s^{ij}$ is the part of the state most relevant to agents $i$ and $j$. For different environments, $f_{\theta_e}$ can be defined manually or obtained through pre-training of binary classification tasks. 
In the context of specific learning, there are several approaches: ($1$) When addressing specific and stable communication-constrained environments, one can estimate the priors of communication links through sampling or empirical data, enabling policy learning to adapt as closely as possible to these specific conditions. 
($2$) In the face of diverse and non-stable communication-constrained environments, designing more diverse priors for communication links can allow policy learning to cover multiple exceptional scenarios. Specifically, message-dropout \cite{kim2019message} is considered a case of this prior modeling, where messages are randomly masked with a certain probability to adapt to communication constraints.
Therefore, incorporating such priors helps distinguish more effectively between lossy and lossless messages.
% 在具体学习时有以下几种方式，在面对特定且稳定的通讯受限环境时，可以通过采样或者经验去估计该通讯链路的先验，使得策略学习尽可能适合该特定情况；在面对多样且非稳定的通讯受限环境时，可以设计更加多样化的该通讯链路的先验，使得策略学习尽可能覆盖多种特殊情况，其中，message-dropout被看作此先验建模的一种情况，即通过随机概率屏蔽message的方式以此适应通讯受限情况。因此，使用该先验可以帮助我们更好的区分有效信息和无效信息

\subsection{Messages' Behavioral Impacts Estimating}
% 基于上一小节内容，接下来一个自然的goal是最大化地利用无损message，同时最小化地避免有损message的影响，其中，最关键的是度量message和behavior的相关性，下面具体我们将通过互信息来表征。
Building on the previous section, a natural goal is to maximize the utilization of lossless messages while minimizing the adverse effects of lossy messages. It is critical to measure the correlation between messages and agent behaviors, where we will specifically characterize via mutual information.
\subsubsection{MI between Messages and Behaviors}
MI is often used to improve multi-agent collaboration in MARL \citep{wang2019learning, li2022pmic}. It is a theory that measures the correlation between different variables. When extended to communication-constrained MARL, a basic requirement is to measure the correlation between messages and agent behaviors. So we have
% 互信息在多智能体强化学习中常被用于改进多智能体协同，是一种度量不同变量间相关性的理论，推广到通讯多智能体强化学习中，一个基本需求便是度量message和智能体行为间的相关性，于是有
\begin{equation}
    I(m^{ji},a^i)=H(a^i)-H(a^i|m^{ji}), j\neq i
\end{equation}
where $H(\cdot)$ and $H(\cdot|\cdot)$ denote the entropy and conditional entropy respectively, and $m^{ji}$ denote the message transmissed from agent $j$ to agent $i$. $H(a^i)$ describes the ability to explore various behaviors of agent $i$, which could help generate diverse trajectories and avoid policy collapse when maximized. $H(a^i|m^{ji})$ measures the behavioral uncertainty of agent $i$, which encourages agent $i$ to behave deterministically given message $m^{ji}$ when minimized.

\subsubsection{Du-MIE for Constrained Communication}
\begin{figure}[!htbp]
	\centering
	\includegraphics[width=1.0\columnwidth]{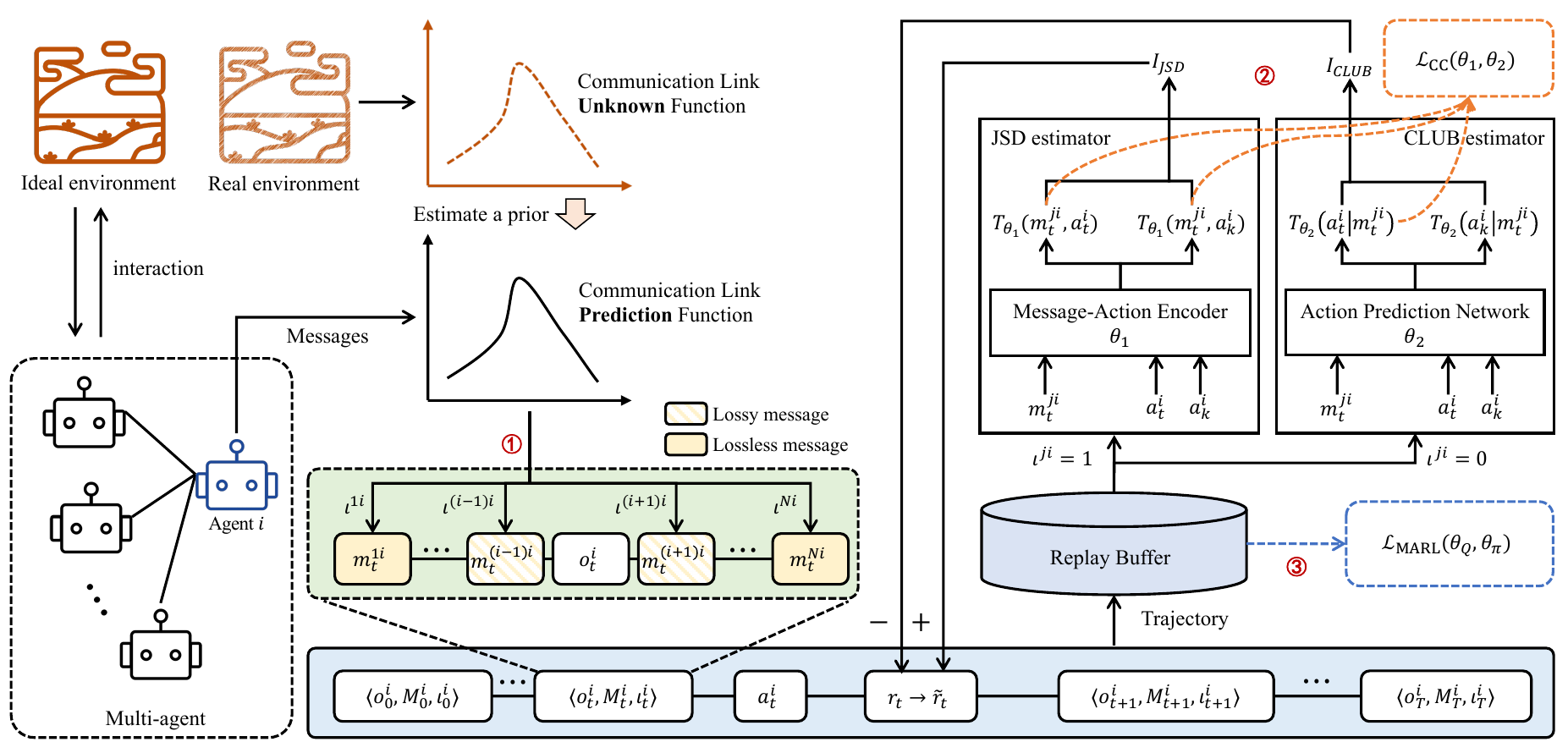}
	\caption{The overall framework for communication-constrained MARL. It can be divided into three main steps: \ding{192} Distinguishing between lossy and lossless messages by constructing communication link priors; \ding{193} Shaping the global reward through learning Du-MIE for constrained communication; \ding{194} Stably optimizing multi-agent policies based on MARL algorithms.}
    % 通讯受限MARL算法的整体框架.主要步骤分为3步：1.通过构建通讯链路先验来区分有效信息和无效信息；2.通过学习Du-MIE for constrained communication, 塑造全局奖励；3.基于MARL算法稳定优化多智能体策略。
	\label{fig_frame}
\end{figure}
% 基于通讯先验和互信息基础定义，在通讯受限下的多智能体强化学习中，我们便可以提出了一个自然的目标，即最大化无损message与智能体行为的互信息（增强无损messages的正相关性），最小化有损message与智能体行为的互信息（去除有损messages的负相关性）。然而计算互信息并非易事，受Dual Mutual Information Estimation启发，我们构造一个面向受限通讯的Du-MIE. 

% 为了实现估计上述目标，我们引入了Dual Mutual Information Neural Estimation，分别基于无损信息和有损信息，通过Jensen-Shannon MI estimator估计MI下界以此来最大化 和 CLUB估计MI上界以此来最小化，具体而言，基于无损信息样本，JSD估计MI下界可以表示为
Note that the basic definitions of communication-constrained prior and MI, we thus formulate a natural objective propose for communication-constrained MARL: maximize the MI between lossless messages and agent behaviors (to enhance their positive correlations) while minimizing the MI between lossy messages and agent behaviors (to mitigate their negative correlations). However, exact computation of MI is highly challenging, as it necessitates simultaneously determining both the joint probability distribution and the marginal distributions of the variables involved \cite{belghazi2018mutual}. 
Inspired by the Dual Mutual Information Estimatior (Du-MIE) \cite{li2022pmic}, we construct a Du-MIE for constrained communication to estimate the above objective.
The Jensen-Shannon MI estimator based on Jensen-Shannon divergence (JSD) \cite{hjelm2019learning} estimates the lower bound of $I(m^{ji}; a^i)$ for maximization and the Contrastive Log-ratio Upper Bound (CLUB) \cite{cheng2020club} estimates the upper bound of $I(m^{ji}; a^i)$ for minimization, based on replay buffer $D$ with lossless messages $\mathcal{M}^+$ and lossy messages $\mathcal{M}^-$ respectively. Specifically, JSD can be defined as
\begin{equation}
    \begin{aligned}
        & I_{\text{JSD}}(m^{ji};a^i)
        = \underbrace{\mathbb{E}_{\mathbb{P}_{\mathcal{MA}^i}}\left[-\text{sp}\left(-T_{\theta_1}(m_t^{ji},a_t^i)\right)\right] 
        -\mathbb{E}_{\mathbb{P}_{\mathcal{M}^i}\otimes \mathbb{P}_{\mathcal{A}^i}}\left[\text{sp}\left(T_{\theta_1}(m_t^{ji}, a_k^i)\right)\right]}_{-\mathcal{L}^{ji}(\theta_1)
        },
    \end{aligned}
\end{equation}
where $\mathbb{P}_{\mathcal{MA}^i}$ is message-action joint distribution for agent $i$, $\mathbb{P}_{\mathcal{M}^i}\otimes \mathbb{P}_{\mathcal{A}^i}$ is the product of the marginals. $\text{sp}(z) = \log(1+e^z)$ is the softplus function. $T_{\theta_1}$ is a discriminator function modeled by a neural network with parameters $\theta_1 \in \Theta$.
$m^{ji}_t$ and $a^i_t$ are obtained by joint sampling from the replay buffer $\mathcal{D}$ with lossless messages $\mathcal{M}^+$, while $a^i_k$ is sampled independently from the same buffer. By minimizing the loss 
$\mathcal{L}^{ji}(\theta_1)$, we can make $I_{\text{JSD}}(m^{ji};a^i)$ closely approximate the lower bound of $I(m^{ji};a^i)$ \cite{hjelm2019learning}.
% 通过最小化损失L可以使得I尽可能逼近互信息的下界
On the contrary, CLUB can be defined as
\begin{equation}
\begin{aligned}
    I_{\text{CLUB}}(m^{ji};a^i)= &        \underbrace{\mathbb{E}_{\mathbb{P}_{\mathcal{MA}^i}}\left[\log T_{\theta_2}(a^i_t|m^{ji}_t)\right]}_{-\mathcal{L}^{ji}(\theta_2)} 
        - \mathbb{E}_{\mathbb{P}_{\mathcal{M}^i}\otimes \mathbb{P}_{\mathcal{A}^i}}\left[\log T_{\theta_2}(a^i_k|m^{ji}_t)\right],
\end{aligned}  
\end{equation}
where $T_{\theta_2}$ is a variational approximation modeled by a neural network with parameters $\theta_2 \in \Theta$. The samples are sampled in the same way as JSD, but are based
on the replay buffer $\mathcal{D}$ with lossy messages $\mathcal{M}^-$. Similarly, by minimizing the loss $\mathcal{L}^{ji}(\theta_2)$, we can make $I_{\text{CLUB}}(m^{ji};a^i)$ closely approximate the upper bound of $I(m^{ji};a^i)$.

Note that the training losses of the JSD and CLUB estimators are $\mathcal{L}^{ji}(\theta_1)$ and $\mathcal{L}^{ji}(\theta_2)$ respectively, then, the overall training loss of Du-MIE for constrained communication is defined as 
\begin{equation}
    \mathcal{L}_{\text{CC}}(\theta_1, \theta_2)=\sum_i \sum_{j\neq i}\iota^{ji}\mathcal{L}^{ji}(\theta_1)+(1-\iota^{ji})\mathcal{L}^{ji}(\theta_2).
    \label{equ:dumie}
\end{equation}

The architecture of Du-MIE for constrained communication are illustrated in \ref{fig_frame}. 
After training, the JSD and CLUB can be used as learning signals to guide the behavior of the agent towards lossless communication and away from lossy communication, which will be introduced in the next subsection.

\subsection{Communication-Constrained MARL}
In order to characterize the impact of the upper and lower bounds of mutual information on multi-agent policy learning, inspired by reward shaping \citep{li2022pmic}, we reshape the global reward as follows:
\begin{equation}
    \tilde{r}_t=r_t+\sum_i \sum_{j\neq i}\alpha\iota^{ji}I_{\text{JSD}}(m^{ji};a^i) - \beta(1-\iota^{ji})I_{\text{CLUB}}(m^{ji};a^i)
    \label{equa:rs}
\end{equation}
where $\alpha, \beta$ are weight coefficients. Then, a new goal in the communication-constrained MALR is to maximize the expected discounted return $\tilde{J}(\pi)=\mathbb{E}_{\pi}\left[\sum_{t=0}^\infty \gamma^t \tilde{r}_t\right]$. 

% 这个reward shaping可以和不同的MARL算法组合，以CTDE-based的marl算法为例，组合的结果展示在算法1
This reward shaping can be combined with different MARL algorithms. Taking CTDE-based MARL algorithms as an example, the results of the combination are presented in Algorithm \ref{alg:ccmaddpg}. 
% 无论是policy-based还是value-based方法，都可以提取一个全局的temporal difference loss，which as follows:
Both policy-based \cite{lowe2017multi} and value-based \cite{rashid2020monotonic} methods can derive a global temporal difference (TD) loss, which is presented as follows:
\begin{equation}
    \begin{aligned}
        \mathcal{L}_{\text{MARL}}(\theta_Q) &=\mathbb{E}_{s_t,a_t,r_t,s_{t+1}\sim \mathcal{D}}\left[\left(Q_{\theta_Q}(s_t,a_t)-\left(\tilde{r}_t+\gamma Q_{\theta_Q^-}(s_{t+1}, a_{t+1})\right)\right)^2\right],
    \end{aligned}
    \label{equa:qvalue}
\end{equation}
where $Q_{\theta_Q}$ is the parameters of the global Q-value network and $Q_{\theta_Q^-}$ is the parameters of the corresponding target network.  For value-based MARL, the joint actions $a_{t+1}$ are obtained via greedy policies $a_{t+1} = \arg\max_{a_{t+1}}Q_{\theta_Q^-}(s_{t+1}, a_{t+1})$. For policy-based MARL, the joint actions $a_{t+1}=\{a^i_{t+1}\}_{i=1}^N$ are obtained via target policies $a_{t+1}^i\sim \pi_{\theta_{\pi^i}^-}(\cdot|o^i_{t+1},M^i_{t+1})$, and an additional policy loss (in actor network) must be incorporated, as follows:
\begin{equation}
    \begin{aligned}
        \mathcal{L}_{\text{MARL}}(\theta_\pi) &= \mathbb{E}_{s_t, a_t \sim \mathcal{D}}\left[-Q_{\theta_Q}(s_t, a_t)\right].
    \end{aligned}
    \label{equa:policy}
\end{equation}
Lines 5–13 present the sampling process of trajectories. Compared to traditional MARL, the samples here not only include additional communication messages and the communication links but also reshape the global reward.
Line 14 shows the Du-MIE training according to the equation (\ref{equ:dumie}). Finally, in line 15, to adapt to different MARL algorithms, the loss function should be correspondingly adjusted, as described in equations (\ref{equa:qvalue}) and (\ref{equa:policy}).
%这奖励塑形可以和不同的通讯多智能体强化学习方法相结合。以带通信的MADDPG为例，该结合方法shown in 算法1. 5-11行呈现了轨迹样本采集过程，相较于传统的MARL，其样本不仅增加了通讯信息和the communication links，同时还reshape奖励信息。第12行表示为根据公式的Du-MIE训练。最后15行为适应不同的MARL算法，其损失函数需要做作对应的调整，正如公式7和8所描述的。
% where $\theta_Q$ and $\theta_{\pi}=\{\theta_{\pi^i}\}_{i=1}^N$ are the parameters of a critic and multiple actors networks respectively. $\theta_Q'$ and $\theta_{\pi}'=\{\theta_{\pi^i}'\}_{i=1}^N$ are the parameters of corresponding target networks.

\begin{algorithm}[!htbp]
   \caption{Communication-Constrained MARL}
   \label{alg:ccmaddpg}
\begin{algorithmic}[1]
   \STATE {\bfseries Input:} maximum episode length $T$, hyperparameters $\alpha$ and $\beta$ to balance the effects of MI, update frequency $k$ for Du-MIE, communication-constrained priors $f_{\theta_{e}}$.
   \STATE {\bfseries Initialize:} main network parameters in MARL $\theta_{Q}, \theta_{\pi}$, corresponding target networks $\theta_{Q}^-$ and $\theta_{\pi}^-$, JSD parameters $\theta_1$, CLUB parameters $\theta_2$.
   % \STATE {\bfseries Initialize:} .
   \STATE {\bfseries Initialize:} experience replay buffer $\mathcal{D}$.
   \REPEAT
   \FOR{$t=1$ {\bfseries to} $T$}
   \STATE Get patial observation $o_t=\{o_t^i\}_{i=1}^N$. 
   \STATE Get message $M_t=\{M_t^i\}_{i=1}^N$ 
   \STATE Predict communication link status $\mathbb{I}_t=\{\iota_t^i\}_{i=1}^N$, $\iota_t^i=\{\iota^{ji}\}_{j\neq i}$.
   \STATE Execute joint actions $a_t=\{a_t^i\}_{i=1}^N$ via sampling $a^i_t\sim\pi^i(\cdot|o_t^i, M_t^i)$.
   \STATE Receive $o_{t+1}=\{o_{t+1}^i\}_{i=1}^N$,  $M_{t+1}=\{M_{t+1}^i\}_{i=1}^N$ and team reward $r_t$.
   \STATE Calculate the shaping reward $\tilde{r}_t$,  according to the equation (\ref{equa:rs}).
   \ENDFOR
   \STATE Store $v=\{o_t, M_t, \mathbb{I}_t, o_{t+1}, M_{t+1}, a_t, \tilde{r}_t\}_{t=1}^T$ to $\mathcal{D}$.
   % \STATE Initialize $noChange = true$.
   % \FOR{$i=1$ {\bfseries to} $m-1$}
   % \IF{$x_i > x_{i+1}$}
   % \STATE Swap $x_i$ and $x_{i+1}$
   % \STATE $noChange = false$
   % \ENDIF
   \STATE Update Du-MIE with replay buffer $\mathcal{D}$ every $k$ steps, according to the equation (\ref{equ:dumie}).
   \STATE Update network parameters in MARL, $\theta_{Q}, \theta_{\pi}$, according to the equations (\ref{equa:qvalue}) and (\ref{equa:policy}).
   \UNTIL{reaching maximum training steps}
\end{algorithmic}
\end{algorithm}

%% file: main/expri.tex
\section{Experiments}
% 这一章节，我们将从整体表现，通讯先验的影响，以及Du-MIE for communication messages的影响三个方面来评估算法的有效性，具体评估内容如下：
% （1）整体性能：主要验证通过我们所提出的算法在不同通讯受限情况下的表现；
% （2）通讯先验的影响：主要验证利用不同通讯先验时所提出方法的表现与性质；
% （3）Du-MIE for communication messages的影响，旨在通过消融性实验，验证该模块对学习通讯受限下多智能体策略的影响。
In this section, we evaluate the algorithm's effectiveness from three aspects: overall performance, the impact of communication-constrained priors (CCPs), and the role of Du-MIE for messages. The specific evaluation contents are as follows: 

\textbf{(1) Overall Performance:} This aims to validate the proposed algorithm’s performance under different communication-constrained scenarios. 

\textbf{(2) Impact of Communication Priors:} This focuses on verifying the performance and properties of the proposed method when using different communication priors. 

\textbf{(3) Role of Du-MIE for Messages:} Through ablation experiments, this evaluation seeks to determine how this module impacts the learning of multi-agent policies under communication constraints.
% \begin{itemize}
%     \item[(1)] Overall Performance: This aims to validate the proposed algorithm’s performance under different communication-constrained scenarios. 
%     \item[(2)] Impact of Communication Priors: This focuses on verifying the performance and properties of the proposed method when using different communication priors. 
%     \item[(3)] Role of Du-MIE for Communication Messages: Through ablation experiments, this evaluation seeks to determine how this module impacts the learning of multi-agent policies under communication constraints.
% \end{itemize}

\subsection{Experimental Setup}
% 在实验的设置中，我们将所提出的算法框架和MADDPG相组合形成communication-constrained MADDPG, 作为验证对象。将其与MAIC算法、全通信的MADDPG算法、带有Dropout的MADDPG算法以及传统的MADDPG算法（即没有通信机制）进行对比。同时，我们选取了MPE作为benchmarks，此外，在通讯受限环境模拟方面，我们分别采用了基于markov模型的和基于距离判定的两种模式，分别去近似地模拟网络丢包情况和诸如水下洞穴等现实一般情况，其具体描述如下：
% （1）markov模型丢包模型：根据状态转移概率矩阵决定所处状态，其中包含一个无损状态和多个有损状态，有损状态越多丢失概率越大，本实验分别设置转移概率矩阵的维度为3，6，8，所对应丢失程度light，medium， heavy；（2）距离判定丢包模式
In the experimental setup, we integrate the proposed algorithm framework with MADDPG \cite{lowe2017multi} to form Communication-Constrained MADDPG (CC-MADDPG) as the primary validation target. It is then compared with four baselines: MAIC \cite{yuan2022multi}, Full-Communication MADDPG (FC-MADDPG), Dropout-MADDPG, and the standard MADDPG, operating without inter-agent communication. We adopt the Multi-Agent Particle Environments (MPEs) \cite{lowe2017multi} as benchmarks. To simulate communication constraints, we employ the following two distinct models,

\textbf{Markov-Based Communication (MBC):} The Markov model \cite{zhang2020succinct} assumes that the state of a system at any time is determined by a state transition probability matrix, which includes one noiseless state and multiple lossy states—the more the number of lossy states, the higher the loss probability. In this experiment, the transition probability matrices are set with dimensions of $3, 6$ and $8$, corresponding to loss levels of light, medium, and heavy, respectively.
% the state of a system at any time depends only on its state at the previous time step, i.e., it exhibits the memoryless property. In the network packet loss model, the network state is defined by whether a packet is lost, with two common states: packet loss state (denoted as State 1) and non-loss state (denoted as State 0). 

\textbf{Distance-Based Communication (DBC):} This approach \cite{yuan2023dacom} simulates signal attenuation in real-world environments (e.g., underwater or cave-like) based on inter-agent distances. Specifically, it sets a distance threshold to determine the degree of communication constraint, where smaller distances lead to higher message loss rates. 
In this experiment, the distance threshold are set with $5, 3$ and $1$, corresponding to loss levels of light, medium and heavy, respectively.
% \begin{itemize}
%     \item Markov Model-Based Communication \cite{zhang2020succinct}: The Markov model assumes that the state of a system at any time depends only on its state at the previous time step, i.e., it exhibits the memoryless property. In the network packet loss model, the network state is defined by whether a packet is lost, with two common states: packet loss state (denoted as State 1) and non-loss state (denoted as State 0). 
%     \item Distance-Based Communication \cite{yuan2023dacom}: This approach simulates signal attenuation in underwater or cave-like environments by dynamically adjusting communication availability based on inter-agent distances. Specifically, it sets a distance threshold to determine the degree of communication constraint, where smaller distances lead to higher message loss rates.
% \end{itemize}

\subsection{Results and Analysis}
% 1. 整体表现，可以适应不同的通讯受限场景，对比message-dropout，最新的通讯方法等，同时在智能体数目较多时，也有不错的效果。
% 2. 通讯先验的影响，如果希望泛化到更多的场景，那么可以找一个多样化的先验，可能鲁棒性能提升，但是最终效果不好；如果是特定场景，可以设计特定先验，能在鲁棒效果之上进一步提升
% 3. 删除MI估计模块，看看效果是否下降，体现MIE的重要性。
\subsubsection{Performance Evaluation}
% 本节旨在通过分析各项算法在不同MPE任务场景及不同通信约束测试环境下的平均回合累计奖励，来系统性地评估CC-MADDPG算法的整体性能和鲁棒性。
This section systematically evaluates the overall performance and robustness of the CC-MADDPG algorithm by analyzing average episode cumulative rewards across different MPE task scenarios and communication-constrained testing environments.

% 表格通过均值和标准差对比了本算法与基线方法在不同任务场景及通信约束下的性能表现。从实验结果可以看出，FC-MADDPG虽然在理想的通信环境下能有不错的表现，但是其性能表现出了对通信质量的高度敏感性：一旦引入任何形式的通信约束，FC-MADDPG的性能就会显著下降。NC-MADDPG在所有测试环境中均表现出最低的性能水平，其平均回合累计奖励值远低于其他使用了通信的算法，这表明在这些MPE任务中智能体间的有效通信对于实现高效协作的重要性。对于Dropout-MADDPG算法，其性能表现随着丢失率的变化而有所不同，总体来说，dropout-0.2的配置能够在多数情况下取得相对更优的效果，而更高的dropout率有时会导致性能的下降，比如在Simple\_Reference场景中，dropout-0.8仅能取得36左右的平均奖励，甚至在大多数测试环境中均显著低于FC-MADDPG，这表明在训练过程中过度增加消息丢失比例反而会损害智能体学习有效协作策略的能力。
Table \ref{tab:comm_performance} compares the performance (mean and standard deviation) of our algorithm with baseline methods across various task scenarios and communication constraints.Experimental results demonstrate that while MADDPG with communication achieves satisfactory performance in ideal communication environments, it exhibits high sensitivity to communication quality. Any form of communication constraint leads to significant performance degradation. MADDPG shows the lowest performance across all testing environments, with its average rewards substantially inferior to other communication-based algorithms, emphasizing the critical role of effective inter-agent communication for efficient collaboration in MPE tasks. For the Dropout-MADDPG algorithm, performance varies with different packet loss rates: the dropout-0.2 configuration generally achieves relatively better results, while higher dropout rates (e.g., 0.8) sometimes cause performance deterioration, as evidenced by its mere 36 average reward in Simple\_Reference scenarios and consistent underperformance compared to FC-MADDPG in most environments, suggesting that excessive message dropout during training undermines agents' ability to learn effective collaborative policies.
\begin{table}[!htbp]
\centering
\caption{Performance Comparison of Multi-Agent Algorithms Under Communication Constraints}
\label{tab:comm_performance}
\footnotesize
\setlength{\tabcolsep}{2pt}
\resizebox{1.0\columnwidth}{!}{
\begin{tabular}{@{}llcccccccc@{}}
\toprule
\multirow{2}{*}{Task Scenario} & \multirow{2}{*}{Testing Environment} & \multicolumn{7}{c}{Algorithms} \\
\cmidrule(lr){3-9}
 & &  \makecell{MAIC} & \makecell{FC-MADDPG} & \multicolumn{3}{c}{\makecell{Dropout-MADDPG}} & \makecell{MADDPG} & \makecell{CC-MADDPG} \\
\cmidrule(lr){5-7}
 & & & 0.2 & 0.5 & 0.8 & & \\
\midrule

\multirow{8}{*}{\shortstack{Simple\_Tag}}
& Unrestricted & 1.5$\pm$4.9 & 75.9$\pm$65.3 & 70.3$\pm$65.9 & 65.9$\pm$62.0 & 72.1$\pm$66.1 & 5.2$\pm$12.8 & \textbf{134.7}$\pm$89.9 \\
& Light MBC (3) & 1.8$\pm$6.0 & 72.9$\pm$65.3 & 70.9$\pm$65.9 & 65.9$\pm$62.1 & 72.5$\pm$66.1 & 5.2$\pm$12.8 & \textbf{133.6}$\pm$89.9 \\
& Medium MBC (6) & 1.6$\pm$8.6 & 67.2$\pm$53.7 & 71.0$\pm$65.7 & 67.1$\pm$62.3 & 72.1$\pm$66.1 & 5.5$\pm$15.6 & \textbf{134.9}$\pm$90.9 \\
& Heavy MBC (8) & 1.7$\pm$6.2 & 54.4$\pm$43.8 & 69.8$\pm$66.5 & 67.9$\pm$63.2 & 72.7$\pm$66.0 & 6.0$\pm$14.2 & \textbf{131.4}$\pm$86.2 \\
& Light DBC (5) & 2.3$\pm$7.8 & 19.5$\pm$38.2 & 69.0$\pm$65.8 & 67.3$\pm$62.4 & 70.8$\pm$64.2 & 23.5$\pm$41.5 & \textbf{136.9}$\pm$89.7 \\
& Medium DBC (3) & 2.4$\pm$6.5 & 10.9$\pm$26.5 & 70.1$\pm$65.8 & 67.8$\pm$62.4 & 70.7$\pm$63.8 & 44.8$\pm$58.4 & \textbf{135.3}$\pm$85.1 \\
& Heavy DBC (1) & 2.7$\pm$9.1 & 1.5$\pm$5.2 & 68.7$\pm$64.3 & 66.3$\pm$60.5 & 71.4$\pm$65.7 & 71.2$\pm$70.9 & \textbf{138.0}$\pm$88.1 \\

% \midrule

% \multirow{7}{*}{\shortstack{Simple\_Tag\\(6 agents)}}
% & unrestricted &  \textbf{138.5} & 84.2 & 78.5 & 73.1 & 6.2 & 131.8 \\
% & Light MBC (3) &  123.9 & 83.5 & 79.5 & 72.9 & 6.2 & \textbf{131.9} \\
% & Medium MBC (6) &  110.8 & 82.1 & 79.6 & 72.0 & 6.4 & \textbf{133.2} \\
% & Heavy MBC (8) &  71.5 & 81.3 & 78.6 & 75.2 & 7.0 & \textbf{131.8} \\
% & Light DBC (5) &  14.7 & 83.0 & 80.1 & 75.7 & 6.6 & \textbf{135.3} \\
% & Medium DBC (3) &  5.4 & 75.5 & 79.3 & 77.2 & 8.0 & \textbf{127.4} \\
% & Heavy DBC (1) &  3.8 & 79.1 & 82.2 & 76.5 & 11.1 & \textbf{122.1} \\

% \midrule

% \multirow{7}{*}{\shortstack{Simple\_Tag\\(9 agents)}}
% & unrestricted &  83.4 & 77.7 & 68.0 & 19.8 & 7.2 & \textbf{86.4} \\
% & Light MBC (3) &  81.2 & 77.8 & 68.1 & 19.5 & 7.3 & \textbf{86.2} \\
% & Medium MBC (6) &  50.3 & 77.0 & 66.3 & 19.0 & 8.3 & \textbf{84.7} \\
% & Heavy MBC (8) &  38.9 & 77.4 & 67.3 & 16.5 & 8.1 & \textbf{87.0} \\
% & Light DBC (5) &  9.7 & 42.9 & 68.9 & 19.9 & 13.9 & 63.1 \\
% & Medium DBC (3) &  8.0 & 27.1 & 66.3 & 20.4 & 18.1 & 50.8 \\
% & Heavy DBC (1) &  5.8 & 21.0 & 69.0 & 21.1 & 17.5 & 39.9 \\

\midrule

\multirow{8}{*}{\shortstack{Simple\_Spread}}
& Unrestricted & -298.0$\pm$75.1 & -138.7$\pm$26.0 & -145.4$\pm$23.5 & -137.5$\pm$27.6 & -138.1$\pm$22.8 & -194.9$\pm$26.7 & \textbf{-129.4}$\pm$20.1 \\
& Light MBC (3) & -293.3$\pm$63.8 & -138.8$\pm$25.5 & -145.1$\pm$23.2 & -138.1$\pm$27.6 & -136.1$\pm$22.7 & -193.5$\pm$26.7 & \textbf{-129.2}$\pm$19.9 \\
& Medium MBC (6) & -295.9$\pm$78.4 & -138.7$\pm$25.4 & -145.4$\pm$22.5 & -138.0$\pm$27.6 & -136.0$\pm$22.5 & -192.9$\pm$27.1 & \textbf{-129.0}$\pm$20.2 \\
& Heavy MBC (8) & -283.7$\pm$51.8 & -142.0$\pm$26.9 & -144.3$\pm$24.2 & -138.1$\pm$27.4 & -135.8$\pm$22.5 & -190.9$\pm$27.7 & \textbf{-128.7}$\pm$19.2 \\
& Light DBC (5) & -301.2$\pm$65.5 & -138.7$\pm$24.4 & -138.8$\pm$23.4 & -145.6$\pm$27.6 & -138.1$\pm$22.0 & -190.5$\pm$24.7 & \textbf{-128.7}$\pm$20.3 \\
& Medium DBC (3) & -282.4$\pm$63.8 & -156.3$\pm$24.9 & -139.7$\pm$21.2 & -143.4$\pm$27.0 & -138.0$\pm$22.9 & -177.4$\pm$25.6 & \textbf{-127.6}$\pm$19.2 \\
& Heavy DBC (1) & -289.5$\pm$63.1 & -191.9$\pm$26.8 & -140.9$\pm$20.8 & -144.5$\pm$27.6 & -138.0$\pm$23.5 & -169.6$\pm$27.1 & \textbf{-128.0}$\pm$20.6 \\

\midrule

\multirow{8}{*}{\shortstack{Simple\_Reference}}
& Unrestricted &  0.2$\pm$1.4 & 51.0$\pm$62.6 & 54.4$\pm$78.3 & 51.6$\pm$69.0 & 38.0$\pm$57.9 & 2.4$\pm$4.1 & \textbf{76.9}$\pm$76.8 \\
& Light MBC (3) &  0.3$\pm$1.7 & 47.2$\pm$62.6 & 54.3$\pm$78.3 & 51.6$\pm$66.4 & 37.3$\pm$58.5 & 2.4$\pm$4.1 & \textbf{76.5}$\pm$76.8 \\
& Medium MBC (6) &  0.2$\pm$1.4 & 42.3$\pm$53.2 & 54.4$\pm$78.0 & 51.8$\pm$67.9 & 38.5$\pm$58.3 & 2.7$\pm$4.0 & \textbf{75.5}$\pm$75.8 \\
& Heavy MBC (8) &  0.2$\pm$1.3 & 27.9$\pm$35.5 & 54.3$\pm$78.2 & 54.9$\pm$67.3 & 36.3$\pm$58.3 & 2.7$\pm$4.3 & \textbf{76.2}$\pm$77.9 \\
& Light DBC (5) &  0.4$\pm$2.0 & 49.3$\pm$62.3 & 53.3$\pm$77.4 & 53.0$\pm$70.3 & 36.0$\pm$58.3 & 23.5$\pm$4.1 & \textbf{75.5}$\pm$76.6 \\
& Medium DBC (3) &  0.3$\pm$1.4 & 39.1$\pm$56.5 & 53.3$\pm$76.4 & 53.3$\pm$71.9 & 37.2$\pm$59.4 & 32.0$\pm$4.2 & \textbf{73.8}$\pm$74.5 \\
& Heavy DBC (1) &  0.1$\pm$1.0 & 4.0$\pm$12.9 & 52.8$\pm$76.5 & 53.2$\pm$70.7 & 34.6$\pm$60.3 & 41.0$\pm$4.1 & \textbf{62.1}$\pm$70.1 \\

\midrule

\multirow{8}{*}{\shortstack{Simple\_Adversary}}
& Unrestricted &  -29.5$\pm$26.4 & -6.7$\pm$5.1 & -6.8$\pm$5.0 & -6.6$\pm$4.8 & -6.7$\pm$4.8 & -75.5$\pm$48.5 & \textbf{-5.8}$\pm$5.0 \\
& Light MBC (3) &  -33.9$\pm$26.7 & -6.7$\pm$5.1 & -6.8$\pm$5.0 & -6.6$\pm$4.8 & -6.7$\pm$4.8 & -75.5$\pm$48.5 & \textbf{-5.9}$\pm$5.0 \\
& Medium MBC (6) &  -35.4$\pm$29.5 & -7.1$\pm$5.0 & -7.2$\pm$5.1 & -6.6$\pm$4.9 & -6.7$\pm$4.8 & -62.9$\pm$41.6 & \textbf{-6.1}$\pm$5.0 \\
& Heavy MBC (8) &  -33.1$\pm$28.7 & -7.5$\pm$5.0 & -6.8$\pm$5.0 & -6.8$\pm$4.9 & -6.7$\pm$4.7 & -51.5$\pm$34.5 & \textbf{-6.1}$\pm$5.1 \\
& Light DBC (5) &  -34.9$\pm$31.0 & -26.6$\pm$15.1 & -10.8$\pm$13.3 & -6.3$\pm$5.0 & -6.7$\pm$4.7 & -14.0$\pm$8.0 & \textbf{-6.4}$\pm$5.8 \\
& Medium DBC (3) &  -29.3$\pm$25.4 & -38.6$\pm$21.3 & -22.4$\pm$19.0 & -6.9$\pm$5.2 & -6.7$\pm$4.7 & -8.8$\pm$5.8 & \textbf{-7.7}$\pm$7.2 \\
& Heavy DBC (1) &  -35.0$\pm$29.3 & -45.3$\pm$21.8 & -27.3$\pm$19.5 & -7.7$\pm$5.3 & -6.8$\pm$4.7 & -7.8$\pm$5.5 & \textbf{-8.8}$\pm$7.4 \\
\bottomrule
\vspace{-2.0em}
\end{tabular}
}
% \vspace{-1.0em}
\end{table}

% CC-MADDPG算法无论是在理想通信环境还是各种模拟的通信约束环境下，取得的平均回合累计奖励值均接近于或者高于其他对比算法，即使是在通信条件极为恶劣的短距离阈值（接近无通信）的测试环境里仍能维持较高的性能，比如在Simple\_Tag(3 agents)场景里仍能维持高达138.0的性能，而此时FC-MADDPG的性能已跌至1.5，证明了CC-MADDPG的有效性与鲁棒性。而与多种通信型基线相比，MAIC 在四个任务上的均值整体处于最低档，且对不同通信约束强度不敏感，这表明MAIC产生的激励消息无法在通讯受限的环境下对协作产生有效引导。
In contrast, CC-MADDPG consistently achieves average rewards comparable to or exceeding other algorithms across both ideal and constrained communication environments. Notably, it maintains superior performance even under extreme communication conditions like heavy distance-based constraint (approaching non-communication scenarios), exemplified by its 138.0 performance in Simple\_Tag when FC-MADDPG deteriorates to 1.5, demonstrating remarkable robustness. Compared with multiple communication-based baselines, MAIC attains the lowest average performance across all four tasks and is largely insensitive to the strength of communication constraints, indicating that the incentive messages it produces fail to effectively guide cooperation in these tasks.

% 此外，一个值得注意的现象是，在本实验设定的时变丢包测试环境（state\_transition）中，无论是轻度、中度还是重度级别的丢包，对于各个算法得到的平均回合累计奖励值影响相对较小，性能表现在不同的网络波动程度下变化不明显。推测可能与以下几个因素有关：首先，实验采用的基于预定义状态转移概率矩阵与马尔可夫模型生成的时变丢包模式，虽然可以模拟无线网络信道丢包的突发性和相关性，但是如果产生的平均丢包率相对比较低，或者丢包的“坏状态”持续时间不够长，不足以持续破坏关键协作时刻，那么可能会被智能体策略的内在鲁棒性所部分抵消；其次，MPE任务中回合长度较短（固定为25个时间步），这可能限制了时变信道特性充分展现其对消息传输的影响。
An interesting observation emerges from our time-varying packet loss environments: performance variations across different network fluctuation levels (light, medium, and heavy) remain relatively insignificant. It may stem from two factors: (1) The predefined Markov-based packet loss patterns, while simulating bursty and correlated wireless channel characteristics, might generate insufficient average loss rates or inadequate durations of "bad states" to persistently disrupt critical collaboration moments, potentially mitigated by inherent robustness in agent policies; (2) The fixed 25-step episode length in MPE tasks may limit full manifestation of time-varying channel impacts.

\subsubsection{Impact of Communication Constraint Priors}
% 本节旨在探讨在多智能体强化学习的训练阶段引入通信约束先验的有效性，并分析不同先验策略对于模型在通信约束环境下性能表现的具体影响。
This section investigates the effectiveness of incorporating communication constraint priors during the multi-agent reinforcement learning training phase, and analyzes how different prior strategies specifically influence model performance in communication-constrained environments.
\begin{table}[!htbp]
\centering
\caption{Performance Comparison of CC-MADDPG with Different Priors}
\label{tab:priors_performance}
\begin{tabular}{@{}llcc@{}}
\toprule
\multirow{2}{*}{\makecell{Task Scenario}} & \multirow{2}{*}{\makecell{Testing Environment}} & \multicolumn{2}{c}{\makecell{Priors Type}} \\
\cmidrule(lr){3-4}
& & \makecell{dropout-0.2} & \makecell{Test-Matched} \\
\midrule

\multirow{3}{*}{\makecell{Simple\_Spread}}
& Light DBC & -128.7$\pm$20.3 & \textbf{-119.0}$\pm$29.4 \\
& Medium DBC & -127.6$\pm$19.2 & \textbf{-98.0}$\pm$30.5 \\
& Heavy DBC & -128.0$\pm$20.6 & \textbf{-107.0}$\pm$33.3 \\

\midrule

\multirow{3}{*}{\makecell{Simple\_Reference}}
& Light DBC & 75.5$\pm$76.6 & \textbf{84.2}$\pm$84.1 \\
& Medium DBC & 73.8$\pm$74.5 & \textbf{107.5}$\pm$109.1 \\
& Heavy DBC & 62.1$\pm$70.1 & \textbf{80.2}$\pm$86.7 \\

\bottomrule
% \vspace{-2.0em}
\end{tabular}

\end{table}

% 通过对比表格\ref{tab:comm_performance}中各算法的表现，可以看出在训练中引入了通信约束先验所带来的优势。以FC-MADDPG作为参照，其虽然在理想测试条件下能够达到较高的性能水平，但是在遭遇任何形式的通信约束时，性能往往会发生灾难性的下降。比如在Simple\_Tag(3 agents)场景中，当测试环境采用基于距离判定的丢包模式时，FC-MADDPG的平均回合累计奖励从理想条件下的75.9分别骤降到19.5、10.9和1.5。这种剧烈的性能衰减充分说明了在理想条件下训练的模型在面对通信约束挑战时的脆弱性。
As evidenced by comparative results in Table \ref{tab:comm_performance}, algorithms introducing communication constraint priors during training demonstrate significant advantages. Using FC-MADDPG as a baseline reference, although it achieves high performance under ideal testing conditions, its performance suffers catastrophic degradation when encountering any form of communication constraints. For instance, in the Simple\_Tag scenario under distance-based constraints, FC-MADDPG's average episode cumulative reward plummets from 75.9 in ideal conditions to 19.5, 10.9, and 1.5 respectively. This drastic performance deterioration highlights the vulnerability of models trained under ideal conditions when facing communication constraints.

% 与此形成鲜明对比的是，那些在训练阶段引入了通信约束先验的算法，比如Dropout-MADDPG和CC-MADDPG，均表现出了更为优越的鲁棒性。具体而言，采用了dropout-0.2先验的Dropout-MADDPG在Simple\_Tag(3 agents)的短距离限制的严苛环境中，其平均回合累计奖励（68.7）远远高于FC-MADDPG（1.5），而CC-MADDPG在同样的dropout-0.2先验基础上，其性能表现（138）甚至超越了其在理想通信环境下的水平。该结果有力地说明了在训练阶段引入通信约束先验确实能够让智能体更好地适应测试阶段的通信约束环境，表现出更好的性能稳定性。
In stark contrast, algorithms incorporating communication constraint priors during training, such as Dropout-MADDPG and CC-MADDPG, exhibit superior robustness. Specifically, Dropout-MADDPG with dropout-0.2 prior achieves an average reward of 68.7 in the heavy distance-based constraint environment of Simple\_Tag, vastly outperforming FC-MADDPG's 1.5. Remarkably, CC-MADDPG utilizing the same dropout-0.2 prior even surpasses its ideal communication environment performance with a score of 138. These results conclusively demonstrate that introducing communication constraint priors during training enables better adaptation to constrained communication environments during testing, yielding enhanced performance stability.

% 为了在未知且多变的测试环境中取得较好的泛化性能，本研究中CC-MADDPG默认采用了dropout-0.2作为标准通信约束先验，这种通用的随机消息丢失先验覆盖了多种潜在的通信约束模式，从而赋予了模型一定的基础鲁棒性。在此基础上，本研究又进一步探讨了一个更深层次的问题：如果训练时采用的通信约束先验能够更精确匹配测试时实际遇到的约束环境，是否能够在已有的鲁棒性基础之上带来额外的性能提升？为了验证这一假设，本研究进行了一组对比试验，表格\ref{tab:priors_performance}展示了CC-MADDPG采用两种不同先验策略时的性能对比。“同测试环境”意味着智能体在训练时经历的通信约束类型与测试时所处的环境完全一致。从表格内容可以看出，采用“同测试环境”作为先验知识训练得到的的模型，其平均回合累计奖励均显著优于采用通用dropout-0.2作为先验知识的模型。比如在Simple\_Spread的中距离限制环境下，通用先验模型的平均奖励值为-127.6，而匹配先验模型的平均奖励值则提升至了-98。以上结果表明，虽然通用的随机消息丢失先验能够赋予模型一定的鲁棒性，但是如果先验知识能够更精确地反映目标部署环境的通信特性，模型的性能可以在此基础上得到进一步的优化。
To ensure satisfactory generalization in unknown and variable testing environments, CC-MADDPG in this study adopts dropout-0.2 as the standard communication constraint prior by default. This generalized random message dropout prior covers multiple potential communication constraint patterns, endowing the model with fundamental robustness. Building on this foundation, we further investigate whether employing training priors that precisely match the actual test environment constraints could yield additional performance improvements. Verification experiments reveal in Table \ref{tab:priors_performance} that models trained with "test-environment-matched" priors (where training constraints exactly match testing constraints) consistently outperform those using generic dropout-0.2 priors. For example, in Simple\_Spread's medium distance-based constraint environment, the generic prior model achieves -127.6 average reward versus -98 for the matched prior model. These findings indicate that while generic message dropout priors provide baseline robustness, precisely tailored priors reflecting target deployment environments can substantially optimize model performance.

\subsubsection{Impact of Dual Mutual Information Optimization Module}
% 本节通过消融实验来深入探究本文引入的双重互信息优化模块的有效性。为了确保对比的公平性，所有进行消融研究的算法变体都使用相同的通信约束先验（dropout-0.2）。
This section investigates the effectiveness of the proposed dual mutual information optimization module through ablation studies. To ensure fairness in comparisons, all algorithm variants in this ablation study employ identical communication constraint priors (dropout-0.2).
% 本研究设计了以下四种算法变体进行对比：
% 基准模型：不使用任何互信息优化，等价于在训练时采用dropout-0.2通信约束先验的MADDPG算法，该基准用于衡量仅依靠先验知识所能达到的性能水平。
% 变体1：仅激活双重互信息优化模块中鼓励利用有效信息的部分。此算法利用JSD互信息估计器来最大化智能体决策动作与有效通信消息之间的互信息下界，该算法变体用于评估单纯增强智能体对有效通信消息的利用所带来的性能提升。
% 变体2：仅激活双重互信息优化模块中抑制无效信息干扰的部分。此算法利用CLUB互信息估计器来最小化智能体决策动作与无效通信消息之间的互信息上界，该算法变体用于评估单纯抑制无效信息对智能体决策的影响所带来的性能提升。
% 完整模型：此设置即为本研究提出的完整算法，同时激活双重互信息优化模块的两个优化方向。
We designed four algorithm variants for systematic comparison:

\textbf{Baseline Model}: Excludes all mutual information optimizations, equivalent to the MADDPG with dropout-0.2 communication constraint prior during training. 
% This serves to measure performance achievable through prior knowledge alone.

\textbf{Variant 1}: Activates only the lossless messages utilization component in Du-MIE module, which employs a JSD-based MIE to maximize the lower bound of mutual information between agent decisions and valid messages.
% , evaluating performance gains from enhanced utilization of effective messages.

\textbf{Variant 2}: Activates only the lossy messages suppression component in the Du-MIE module, which only utilizes a CLUB-based MIE to minimize the upper bound of mutual information between agent decisions and invalid messages.
% , assessing performance improvements from suppressing detrimental message interference.

\textbf{Full Model}: Represents our complete proposed algorithm with both optimization directions activated in Du-MIE for messages.

Table \ref{tab:mi_ablation} summarizes the average episode cumulative rewards of these four variants across various communication-constrained testing environments in Simple\_Tag and Simple\_Spread scenarios. 
Between the baseline and single-component variants, both mutual information optimizations independently enhance performance. In Simple\_Tag, the baseline achieves ~70 average rewards across environments. Variant 1 (maximization-only) elevates rewards to ~81, while Variant 2 (minimization-only) significantly boosts performance to ~120. This confirms the intrinsic value of mutual information optimization beyond prior knowledge utilization.

The full model consistently outperforms all variants, demonstrating synergistic complementarity between the two optimization components. For instance, in Simple\_Tag's most constrained short-distance environment, the full model achieves 138.0 rewards versus 68.7 (baseline), 81.7 (Variant 1), and 120.4 (Variant 2). Similar patterns emerge in Simple\_Spread, where the full model (-127.6 to -128.0) substantially surpasses the baseline (-143.4 to -145.6). These results validate that dual-directional mutual information optimization synergistically enhances multi-agent collaboration and system robustness under communication constraints.

\begin{table}[!htbp]
\centering
\caption{Ablation Study on Dual Mutual Information Module}
\label{tab:mi_ablation}
\footnotesize
\setlength{\tabcolsep}{2pt}
\resizebox{1.\columnwidth}{!}{
\begin{tabular}{@{} lccccccccc @{}}
\toprule
\multirow{2}{*}{Scenario} & \multicolumn{2}{c}{MI Coefficients} & \multicolumn{7}{c}{Average Episode Cumulative Reward} \\
\cmidrule(lr){2-3} \cmidrule(lr){4-10}
& {min} & {max} & {Unrestricted} & {Light MBC} & {Medium MBC} & {Heavy MBC} & {Light DBC} & {Medium DBC} & {Heavy DBC} \\
\midrule

\multirow{4}{*}{\makecell{Simple\_Tag}}
& 0 & 0 & 70.3$\pm$65.9 & 70.9$\pm$65.9 & 71.0$\pm$65.7 & 69.8$\pm$66.5 & 69.0$\pm$65.8 & 70.1$\pm$65.8 & 68.7$\pm$64.3 \\
& 0 & 0.01 & 81.3$\pm$75.2 & 81.3$\pm$75.2 & 81.7$\pm$77.0 & 81.6$\pm$74.0 & 81.8$\pm$74.5 & 81.7$\pm$75.6 & 81.7$\pm$73.3 \\
& 0.001 & 0 & 120.5$\pm$82.1 & 119.8$\pm$81.0 & 120.3$\pm$81.8 & 119.3$\pm$78.3 & 121.5$\pm$84.5 & 119.5$\pm$80.5 & 120.4$\pm$78.7 \\
& 0.001 & 0.01 & \textbf{134.7}$\pm$89.9 & \textbf{133.6}$\pm$89.9 & $\textbf{134.9}$$\pm$90.9 & $\textbf{131.4}$$\pm$86.2 & $\textbf{136.9}$$\pm$89.7 & $\textbf{135.3}$$\pm$85.1 & $\textbf{138.0}$$\pm$88.1 \\

\midrule

\multirow{4}{*}{\makecell{Simple\_Spread}}
& 0 & 0 & -145.4$\pm$23.5 & -145.1$\pm$23.2 & -145.4$\pm$22.5 & -144.3$\pm$24.2 & -145.6$\pm$23.4 & -143.4$\pm$21.2 & -144.5$\pm$20.8 \\
& 0 & 0.01 & -139.1$\pm$24.8 & -139.1$\pm$24.8 & -138.7$\pm$24.6 & -138.7$\pm$25.2 & -138.8$\pm$24.4 & -139.7$\pm$23.9 & -140.0$\pm$24.0 \\
& 0.01 & 0 & -143.0$\pm$24.8 & -142.5$\pm$24.8 & -142.0$\pm$27.1 & -141.9$\pm$25.4 & -142.4$\pm$24.8 & -141.7$\pm$26.8 & -139.8$\pm$24.6 \\
& 0.01 & 0.01 & \textbf{-129.4}$\pm$20.1 & $\textbf{-129.2}$$\pm$19.9 & $\textbf{-129.0}$$\pm$20.2 & $\textbf{-128.7}$$\pm$19.2 & $\textbf{-128.7}$$\pm$20.3 & $\textbf{-127.6}$$\pm$19.2 & $\textbf{-128.0}$$\pm$20.6 \\

\bottomrule
% \vspace{-2.0em}
\end{tabular}
}
\end{table}

%% file: main/conc.tex
\section{Conclusion}
Lossy communication remains a critical barrier hindering the practical deployment of MARL in real-world scenarios. To address this challenge, we propose a novel communication-constrained MARL framework that first establishes a unified prior over communication constraints via systematic modeling of lossy communication patterns, enabling agents to adapt strategies across diverse communication-constrained scenarios. Second, by distinguishing between lossy and lossless messages, we develop the Du-MIE to quantify the impact of messages on agent behavior, integrating this into the reward function to enhance the positive influence of reliable messages and mitigate the negative effects of corrupted messages. Finally, when integrated with the MADDPG algorithm, our approach demonstrates superior performance in both overall tasks and ablation studies across benchmark environments, validating its effectiveness in maintaining robust cooperative decision-making under varying communication constraints.

% 在将来，仍然有一下几点值得进一步研究：（1）算法可扩展性：是否能够扩展到Value-based的学习框架；（2）动态环境适应性：在异常动态的通讯受限环境下能否自适应学习鲁棒策略？
In the future, several directions warrant further investigation:
(1) Algorithmic Scalability: Whether the framework can be extended to Value-based learning frameworks;
(2) Adaptability to Dynamic Environments: Whether it can adaptively learn robust policies in highly dynamic communication-constrained environments?